\documentclass{article}
\usepackage{spconf,amsmath,graphicx,amssymb,subcaption,booktabs, multirow,amsthm}

\newcommand{\bs}[1]{\boldsymbol{#1}}
\newcommand{\R}{\mathbb{R}}

\newcommand{\W}{\boldsymbol{W}}
\newcommand{\V}{\boldsymbol{V}}
\newcommand{\x}{\boldsymbol{x}}
\newcommand{\y}{\boldsymbol{y}}
\newcommand{\z}{\boldsymbol{z}}
\renewcommand{\H}{\boldsymbol{H}}
\newcommand{\J}{\boldsymbol{J}}
\renewcommand{\b}{\boldsymbol{b}}

\newcommand{\cmgn}{\texttt{C-MGN}}
\newcommand{\mmgn}{\texttt{M-MGN}}
\newtheorem{prop}{Proposition}

\title{Learning Gradients of Convex Functions with Monotone Gradient Networks}
\name{Shreyas Chaudhari* \qquad Srinivasa Pranav* \qquad Jos\'e M.F. Moura \thanks{* Authors contributed equally.
Authors partially supported by NSF Graduate Research Fellowships (GRFP; Grants DGE1745016, DGE2140739)
and XSEDE \cite{6866038}
Allocation ELE220003
on PSC Bridges-2 system \cite{10.1145/3437359.3465593}.
S. Pranav partially supported by an ARCS Fellowship. 
Authors thank Tyler Vuong for insightful discussions.
}}
\address{Electrical and Computer Engineering, Carnegie Mellon University}
\begin{document}
%
\maketitle
\begin{abstract}
While much effort has been devoted to deriving and analyzing effective convex formulations of signal processing problems, the gradients of convex functions also have critical applications ranging from gradient-based optimization to optimal transport. Recent works have explored data-driven methods for learning convex objective functions, but learning their monotone gradients is seldom studied. In this work, we propose \cmgn\ and \mmgn, two monotone gradient neural network architectures for directly learning the gradients of convex functions. We show that, compared to state of the art methods, our networks are easier to train, learn monotone gradient fields more accurately, and use significantly fewer parameters. We further demonstrate their ability to learn optimal transport mappings to augment driving image data.
\end{abstract}
\section{Introduction}
\label{sec:intro}

Convex functions have been studied and celebrated for their amenable analytic properties, relative ease of optimization, and plethora of applications. When finding solutions to signal processing problems, convex formulations enable us to easily augment objective functions and incorporate prior domain knowledge regarding the structure of the solution. However, for complex problems where prior domain knowledge is either lacking or insufficient, deep learning approaches are attractive alternatives that rely on purely data-driven, nonconvex, and overparameterized problem formulations. Deep neural networks have achieved state of the art performance on a variety of image and speech processing tasks at the cost of sacrificing many benefits of convex optimization: computational efficiency, interpretability, and theoretical guarantees. Thus, even in the age of deep learning, convex optimization methods offer significant value.


Formulating convex optimization problems is an active area of research that permeates nearly all of signal processing, from source localization in communications to image deblurring \cite{palomar2010convex}. However, it is often a laborious process that involves manually designing suitable convex objectives and associated convex constraints. Perhaps more important than the objective function itself is the \textit{gradient} of the function, since most convex problems are solved using computationally frugal gradient-based methods. Monotone gradient maps of convex functions also have critical applications in domains including gradient-based optimization, generalized linear models, linear inverse problems, and optimal transport. Therefore, in this work, we propose to \textit{learn} the gradient of convex functions in a data-driven manner using deep learning. Our approach is a fundamental step toward blending strengths of both deep learning and convex optimization and offers a wide array of applications in data science and signal processing.

\textbf{Contributions}: We propose two neural network architectures for learning gradients of convex functions, i.e., monotone gradient functions \cite{rockafellar1970convex}. To the best of our knowledge, we are the first to propose a method for \textit{directly} parameterizing and learning monotone gradients of convex functions, without first learning the underlying convex function or its Hessian. In contrast to current methods, our networks are considerably easier to train and generalize to high-dimensional problem settings. In this work, we show empirically the efficacy of our approach on a set of standard problems and an image color domain adaptation task.
\section{Related Work}
\subsection{Learning Loss Functions and Regularizers}
Parameterized, monotone gradient functions are useful when we want to optimize input data to minimize a desired loss function that is difficult to express analytically.
Recent works explored \textit{learning} parameters of an objective function during training and then using it to optimize an input at inference time. The adversarial method for training a regularizer in \cite{AdversarialRegularizer} entails a nonconvex optimization problem to generate predictions at inference time. In contrast, an Input Convex Neural Network (ICNN) \cite{ICNN} constrains the learned objective function to be convex with respect to its input. To optimize a proposed input using gradient descent updates, the ICNN must be differentiated at inference time. In this work, we avoid the extra differentiation step by directly learning the gradient of the convex objective. Furthermore, ICNNs are infamously hard to train \cite{ICGN} as most weights must be positive and nonlinearities must be convex, monotonically increasing  functions. 


\subsection{Monotone Gradient Solutions to Optimal Transport Problems}
Directly learning the monotone gradient of a convex function is fruitful even when the convex function itself is unknown. Monotone gradients are significant in optimal transport (Monge) problems that arise in numerous scenarios, including domain adaptation \cite{flamary2016optimal, redko2019optimal}. For example, computer vision algorithms used in autonomous vehicle applications require large amounts of human-annotated training images that may be too expensive or infeasible to obtain for a wide variety of factors like location, lighting, and weather conditions. One promising generative approach to augmenting training data involves solving an optimal transport problem to transform existing annotated road images into similar images under different visual settings. For example, daytime images can be mapped to realistic twilight images of the same scene by solving an optimal transport problem between estimated daytime and nighttime image distributions. Formally, one can solve
\begin{align}
    \inf_{g:g(x)\sim p_Y} \mathbb{E}_{x\sim p_X}\left[c(x,g(x))\right]
    \label{eqn:monge}
\end{align}
for a given cost function $c$ in order to determine an invertible mapping function $g$ that transforms a random variable $x \sim p_X$ to $y \sim p_Y$. Choosing the cost function to be squared Euclidean distance, Brenier's theorem \cite{brenier1991polar, santambrogio2015optimal} states that the unique optimal mapping function $g$ that solves Eqn.~\ref{eqn:monge} is a monotone gradient of a convex function. Brenier's theorem inspired several recent normalizing flow works \cite{ConvexPotentialFlows, makkuva2020optimal} which attempt to optimally transport one probability distribution to another using the gradient of a learnable convex function. These methods train an ICNN \cite{ICNN} for the sole purpose of extracting its gradient map using automatic differentiation. These approaches avoid directly parameterizing the gradient map, are computationally inefficient, and are subject to the difficulties of training ICNNs. 

Input Convex Gradient Networks (ICGNs) \cite{ICGN} take an indirect and more complex approach to alleviate the issues associated with ICNNs: they use a standard neural network to learn factors of a positive semidefinite Hessian matrix and numerically approximate its integral. This computationally expensive approach is not feasible for high-dimensional problems and requires that the neural network always satisfies a partial differential equation. Currently, only zero hidden layer perceptrons are known to always satisfy this condition. While related theoretical analysis is sparse, the symmetry of a scalar-valued function's second derivatives reveals that approximating the gradient of any function using a standard multilayer perceptron with more than one hidden layer requires only representing one feature in the first hidden layer \cite{GradientNetworks}. In this work, we propose two specific neural network architectures designed with explicit weight-tying that avoid this limitation.

\section{Problem Statement}
\label{sec:background}

We learn a monotone gradient function $g(\x) : \R^n \to \R^n$ such that $g(\x) = \nabla f(\x)$, for some convex, twice differentiable function $f(\x) : \R^n \to \R$. A differentiable function $f(\x)$ is convex if and only if its gradient $g(\x)$ is monotone \cite{boyd2004convex}:
\begin{align}
\langle g(\x) - g(\y), \x - \y \rangle \geq 0 \;\;\forall \x,\y \in \R^n
\label{eqn:monotonicity}
\end{align}
Since this condition is difficult to enforce in practice, we rely on
a twice differentiable function $f(\x)$ being convex if and only if its Hessian is positive semidefinite (PSD) \cite{boyd2004convex}:
\begin{align}
    \H_f(\x) = \J_g(\x) \succeq 0 \;\;\forall \x \in \R^n
\end{align}

\noindent This condition subsumes the requirement of a symmetric Jacobian $\J_g$. Thus, parameterizing $g(\x)$ with a neural network requires the Jacobian of the neural network to be PSD everywhere. \textbf{Note:} in this work, Jacobians of neural networks are always computed with respect to their \textit{inputs}.
\section{Proposed Approaches}
\label{sec:proposed_approach}
We propose two monotone gradient neural network (MGN) architectures to learn gradients of convex functions: \cmgn\; and \mmgn\;. They are motivated by two different approaches that ensure a neural network's Jacobian (with respect to its input) is PSD everywhere.
\subsection{Cascaded Network: \cmgn}
We propose a \textit{cascaded} monotone gradient network (\cmgn):
\begin{align}
    \z_0 &= \W\x+\b_0\\
    \z_\ell &= \W\x + \sigma_\ell(\z_{\ell-1}) + \b_\ell\\
    \text{\cmgn}(\x) &= \W^\top\sigma_L\left(\z_{L-1}\right) + \V^\top\V\x + \b_{L}
    \label{eqn:cmgn}
\end{align}
Layer outputs $\z_{\ell}$, biases $\b_{\ell}$, and activation functions $\sigma_\ell$ can differ for each layer $\ell$, but all $L$ layers share weight matrix $\W$. Prop.~\ref{prop:cmgn} shows that the proposed \cmgn\ is monotone for popular activation functions, e.g., tanh, sigmoid, and softplus. 
\begin{prop}
\label{prop:cmgn}
If $\forall \ell \;\sigma_\ell(\cdot)$ is a differentiable, element-wise activation function that is monotonically increasing, then $\forall \x\; \J_{\text{\cmgn}}(\x)\succeq 0$.
\end{prop} 
\begin{proof}
The Jacobian of $\cmgn(\x)$ with respect to $\x$ is:
\begin{align}
 \J_{\text{\cmgn}}(\x)
    &= \W^\top \left(\sum_{\ell=1}^{L}\prod_{i=\ell}^{L} \J_{\sigma_i}\left(\z_{i-1}\right)\right) \W + \V^\top\V 
\end{align}
If each $\sigma_i(\cdot)$ is a monotonically increasing, element-wise function, then each $\J_{\sigma_i}$ is diagonal and non-negative. Therefore, $\sum_{\ell=1}^{L}\prod_{i=\ell}^{L}\J_{\sigma_i}(\z_{i-1})$ is a diagonal, non-negative matrix and is PSD. Since $\W^\top \bs{A} \W \succeq 0$ for any $\bs{A} \succeq 0$ and the sum of PSD matrices is also PSD, $\forall \x\; \J_{\text{\cmgn}}(\x) \succeq 0$.
\end{proof}

\subsection{Modular Network: \mmgn}
We define a modular monotone gradient network (\mmgn):
\begin{align}
\label{eqn:m-mgn}
    \z_k &= \W_k\x + \b_k\\
    \text{\mmgn}(\x) &= \bs{a} + \V^\top\V\x + \sum_{k=1}^K s_k\left(\z_k \right)\W_k^\top\sigma_k\left(\z_k\right)
\end{align}
where $\bs{a}$ is a bias parameter and the number of modules $K$ can be tuned based on the application. The activation function $\sigma_k$, weight matrix $\W_k$, and bias $\b_k$ can differ for each network module. Prop.~\ref{prop:mmgn} relates scalar-valued functions $s_k$ to $\sigma_k$.

\begin{prop}
\label{prop:mmgn}
If $\forall k\ s_k(\x)$ is a convex, twice differentiable, nonnegative scalar function and $\sigma_k(\cdot) = \nabla s_k(\cdot)$, then $\forall \x\; \J_{\mmgn}(\x) \succeq 0$.
\end{prop}

\begin{proof}
The Jacobian of $\text{\mmgn}(\x)$ with respect to $\x$ is:
\begin{align}
  \J_\text{\mmgn}(\x) &= \V^\top\V + \sum_{k=1}^K \Bigg[ s_k\left(\z_k\right) \W_k^\top\J_{\sigma_k}(\z_k)\W_k \nonumber\\
  &\qquad+\left(\W_k^\top\sigma_k(\z_k)\right)\left(\W_k^\top\sigma_k(\z_k)\right)^\top \Bigg]
\end{align}

\noindent Observe that convex $s_k(\cdot)$ implies $\J_{\sigma_k}(\x)\succeq 0$. Since $s_k(\cdot)$ is a scalar, nonnegative function and $\bs{A} \succeq 0$ implies $\bs{B}^\top\bs{A}\bs{B} \succeq 0$, the first term in the summation is PSD. The second term in the summation and $\V^\top \V$ are PSD as they are Gram matrices. Sums of PSD matrices are PSD, therefore $\J_{\mmgn} \succeq 0$.
\end{proof}
\noindent Note that  requiring $s_k(\x) \geq 0$ is not restrictive. For example, if $\x_i$ denotes the $i^{\text{th}}$ component of $\x$, then $s_k(\x) = \log\cosh(\x_1) + \dots + \log\cosh(\x_n)$ yields the element-wise activation $\sigma_k(\x)=\tanh(\x)$. Similar functions $s_k(\cdot)$ can be easily found when $\sigma_k(\cdot)$ is the element-wise softplus or sigmoid functions. Furthermore, the activations $\sigma_k(\cdot)$ in the \mmgn\ are not restricted to be element-wise.

\subsection{Discussion}
The main distinction between the aforementioned approaches is that is that \cmgn\; represents monotone gradients via a single \textit{deep} network and \mmgn\; via several \textit{parallel} networks. Additionally, note that both $\cmgn$ and $\mmgn$ can easily be generalized to incorporate convolutional layers. The $\W$ in \cmgn\; and each $\W_k$ in \mmgn\; can be replaced with any linear operator while maintaining a PSD Jacobian -- assuming the transposed weight matrices are replaced with the adjoint operator. Furthermore, the network outputs can be easily adjusted to represent gradients of strongly convex functions. A twice differentiable function $f(\x)$ is \textit{strongly} convex if there exists $\gamma > 0$ such that $\H_f(\x) \succeq \gamma\bs{I}$. By adding a scalar multiple of the input to the output, we can ensure that the learned gradients correspond to strongly convex functions and that the network is invertible \cite{ rockafellar1970convex,ConvexPotentialFlows}. In particular, this allows us to use our network architecture as a normalizing flow.


Since non-negative diagonal matrices are closed under matrix multiplication, \cmgn's Jacobian would remain PSD even if different learnable, non-negative diagonal matrices were composed after (to the left of) each $\W$ and $\sigma_\ell(\cdot)$. The same applies to \mmgn\ modules where the activations $\sigma_k(\cdot)$ operate element-wise. In practice, we observe that these non-negative diagonal matrices (which scale dimensions) avoid training difficulties faced by non-negative dense matrices in ICNNs (which perform conical combinations) \cite{ICNN}.

\section{Experiments}
\label{sec:experiments}

\subsection{Gradient Field}
\begin{figure}[htbp]
    \centering
    \ninept
    \includegraphics[width=\linewidth, trim={2.5cm .25cm 0cm 0.5cm},clip]{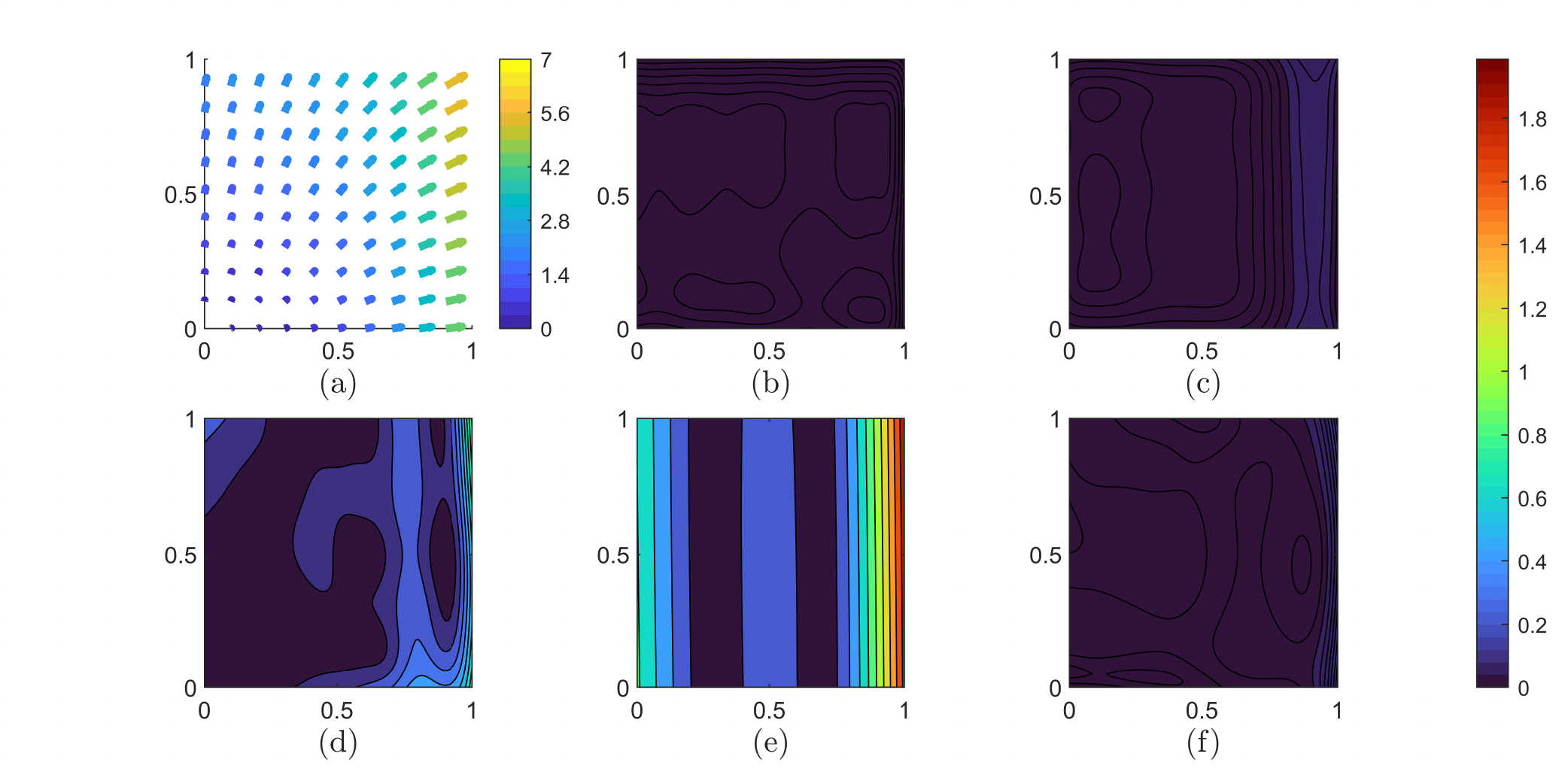}
    \caption{(a) Gradient of $f(\x)$ (Eqn.~\ref{eqn:gradfield_func}) - arrow color corresponds to $\ell_2$ norm of $\nabla f(\x)$. $\ell_2$-error maps between true and predicted gradient for: (b) \cmgn* (c) \mmgn* \;(d) ICGN (e) ICNN with 78 parameters (f) ICNN with 163 parameters}
    \label{fig:gradfield_results}
\end{figure}

We compare our proposed models to ICGNs \cite{ICGN} and ICNNs \cite{ICNN} using the standard problem in \cite{ICGN}: estimating the gradient field of $f(\x)$ over the unit square, where $\x = \begin{bmatrix} \x_1 \;\;  \x_2\end{bmatrix}^\top$ and
\begin{align}
    f(\x) &= \x_1^4 + \frac{\x_2}{2} + \frac{\x_1\x_2}{2} + \frac{3\x_2^2}{2} - \frac{\x_2^3}{3}
    \label{eqn:gradfield_func}
\end{align}
We adopt the procedure used in \cite{ICGN} and train all models using 1 million points randomly sampled from the unit square and mean absolute error loss. All models were trained using the same batch size, learning rate, and number of epochs. Fig.~\ref{fig:gradfield_results} illustrates the gradient field of $f(\x)$ and the $\ell_2$-norm errors between true and predicted gradients. We note that our proposed \cmgn \; and \mmgn \; more accurately learn $\nabla f(\x)$ while requiring considerably fewer parameters.

\begin{table}[ht]
\centering
\ninept
\caption{Results for estimating $\nabla f(\x)$ from Eq.~\ref{eqn:gradfield_func}. *Our methods.}
\begin{tabular}{@{}cccccc@{}}
\cmidrule(l){2-6}
           & ICGN \cite{ICGN} & \multicolumn{2}{c}{ICNN \cite{ICNN}} & \cmgn* & \mmgn* \\ \cmidrule(l){2-6} 
Parameters &  15    &  78    &  163    &  14    & 22     \\
MSE (dB) &    -15.00    &  -4.15    & -30.88   & -39.10  &    -32.31       \\ \cmidrule(l){2-6} 
\end{tabular}
\label{tab:gradfield_results}
\end{table}

\noindent As shown in Table~\ref{tab:gradfield_results}, our proposed \cmgn\; achieves a $\approx$10 dB lower MSE than an ICNN while using less than 10\% of the number of parameters.

\subsection{Optimal Coupling}
\begin{figure}[htbp]
\ninept
\centering
\begin{subfigure}{.32\linewidth}
  \centering
  \includegraphics[width=0.75\linewidth]{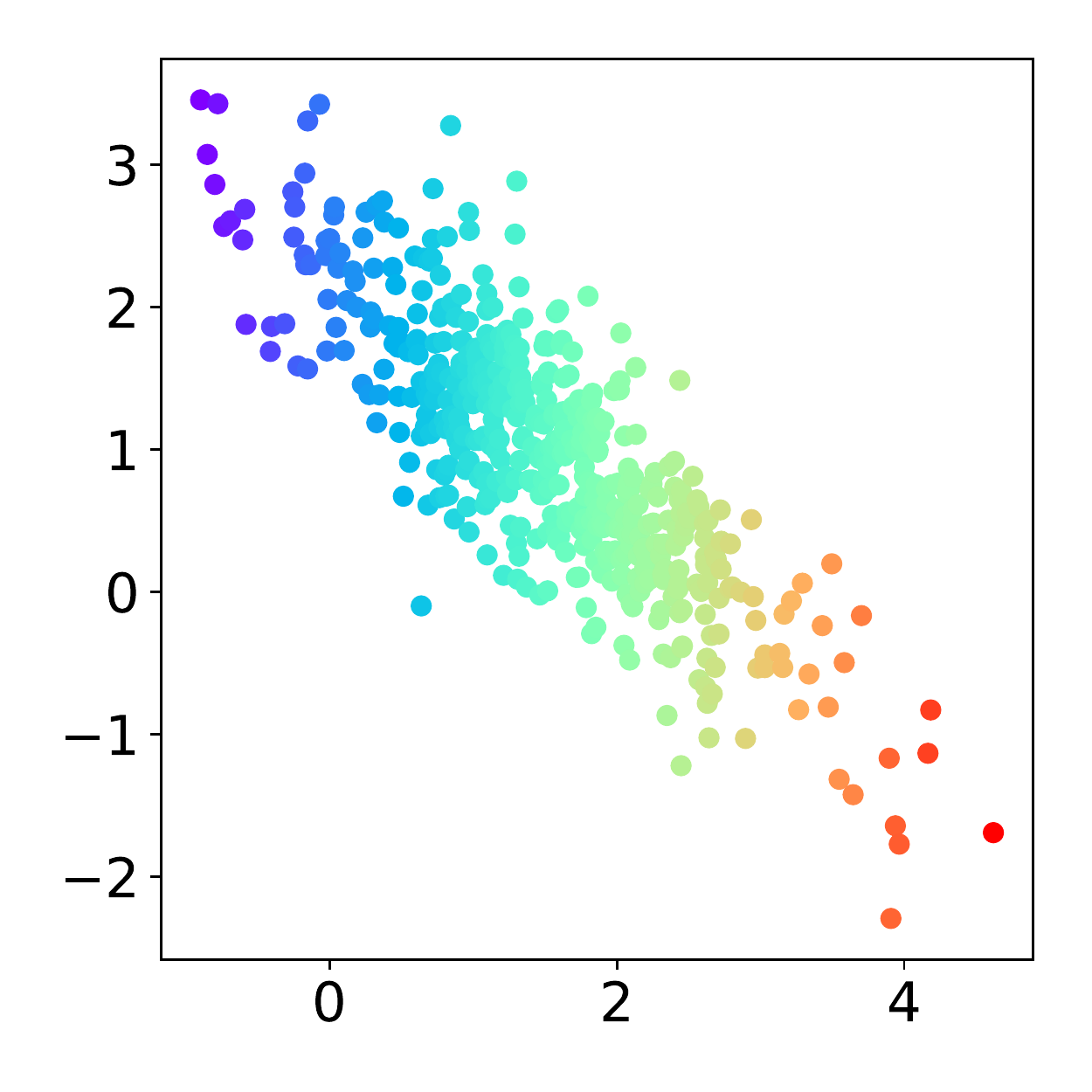}
  \caption{}
  \label{fig:sub-first2}
\end{subfigure}
\begin{subfigure}{.32\linewidth}
  \centering
  \includegraphics[width=0.75\linewidth]{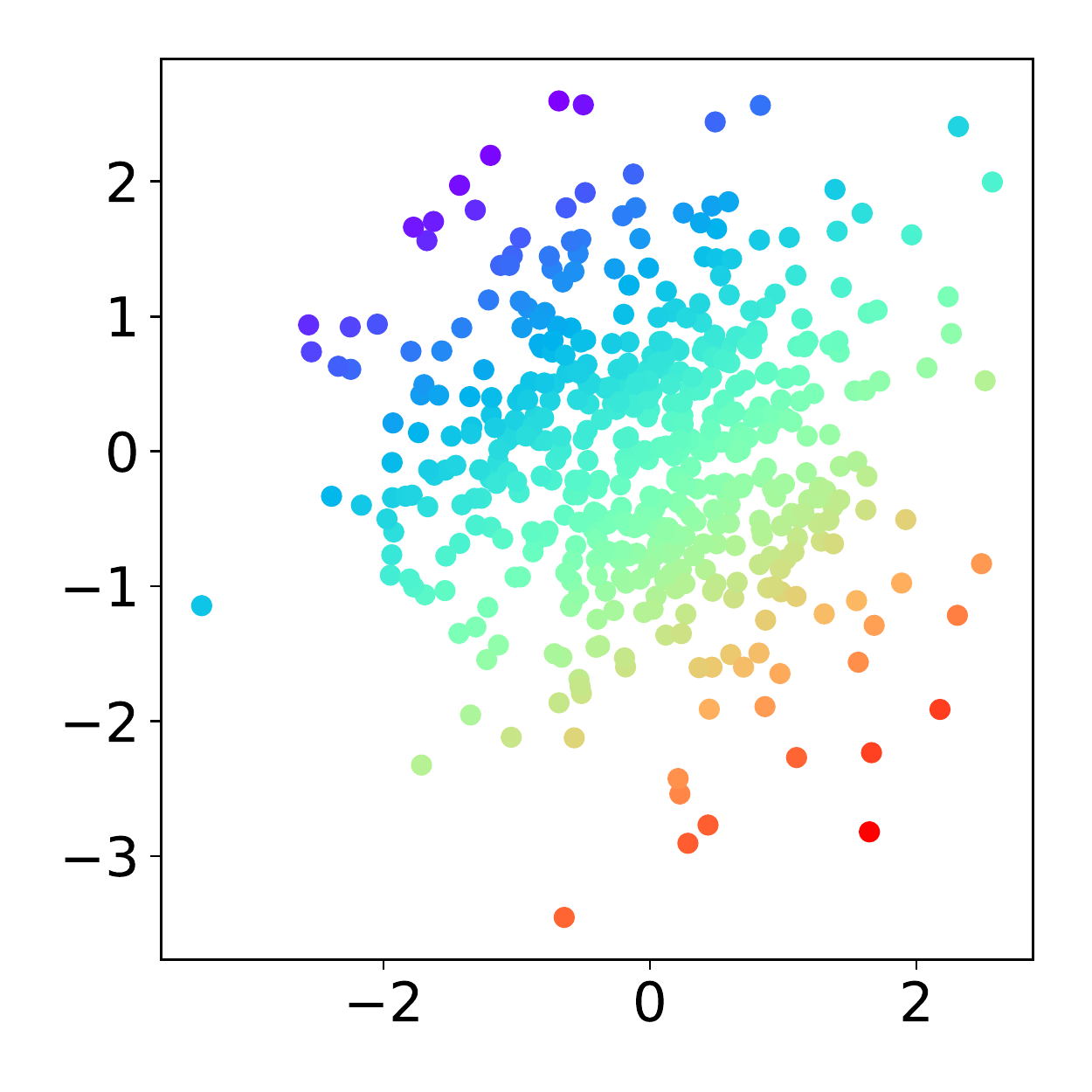}
  \caption{}
  \label{fig:sub-second2}
\end{subfigure}
\begin{subfigure}{.32\linewidth}
  \centering
  \includegraphics[width=0.75\linewidth]{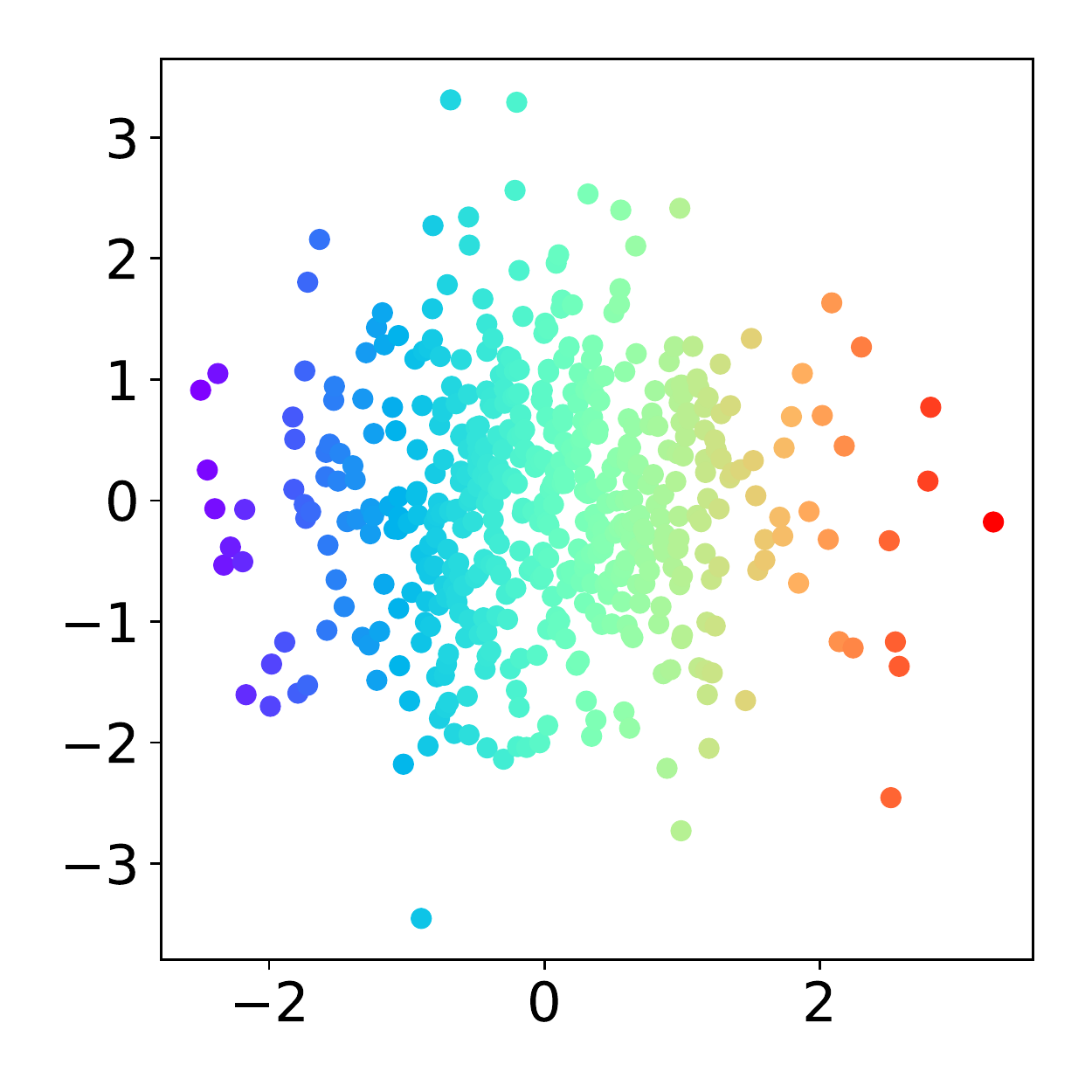} 
  \caption{}
  \label{fig:sub-third}
\end{subfigure}
\newline
\begin{subfigure}{.32\linewidth}
  \centering
  \includegraphics[width=0.75\linewidth]{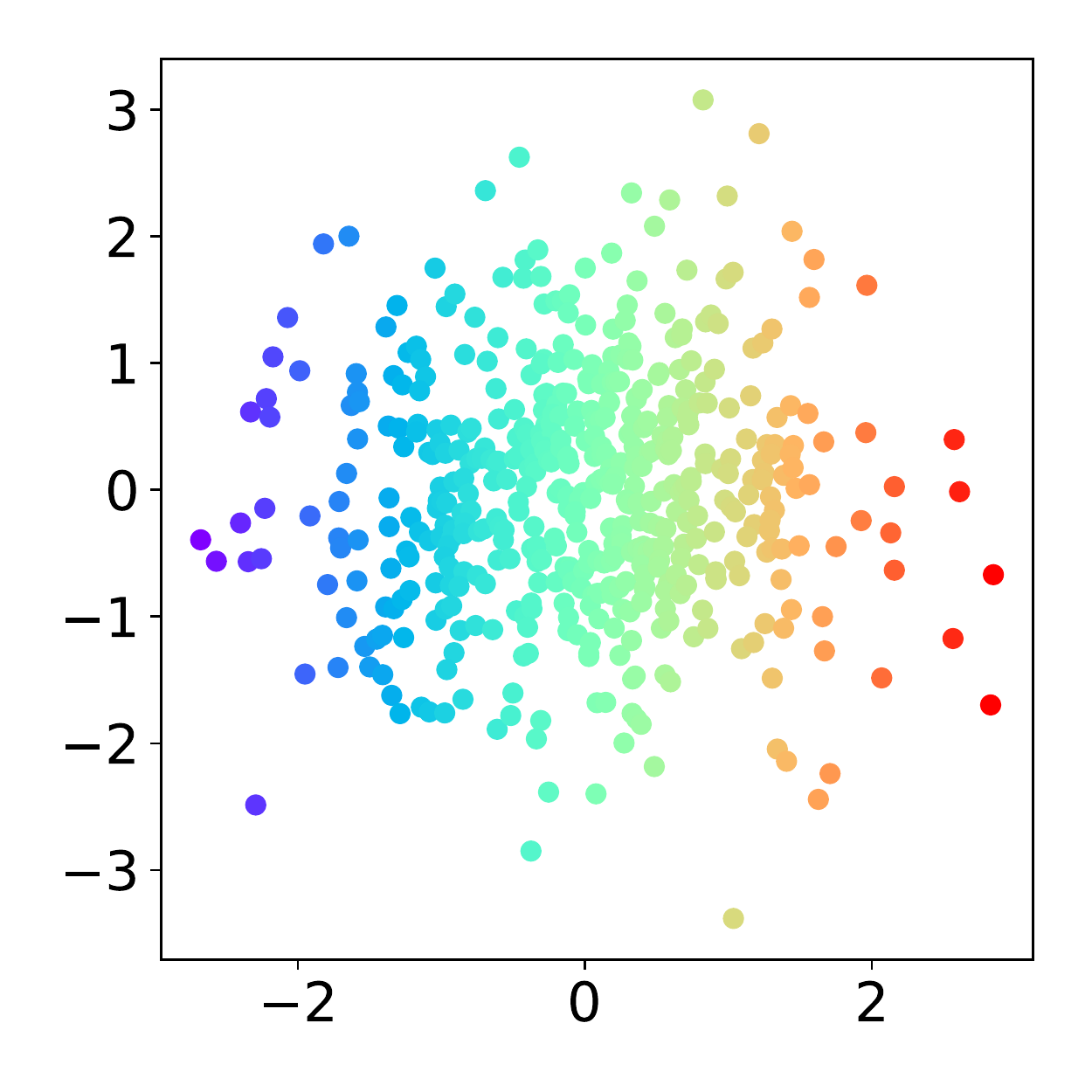}
  \caption{}
  \label{fig:sub-fourth}
\end{subfigure}
\begin{subfigure}{.32\linewidth}
  \centering
  \includegraphics[width=0.75\linewidth]{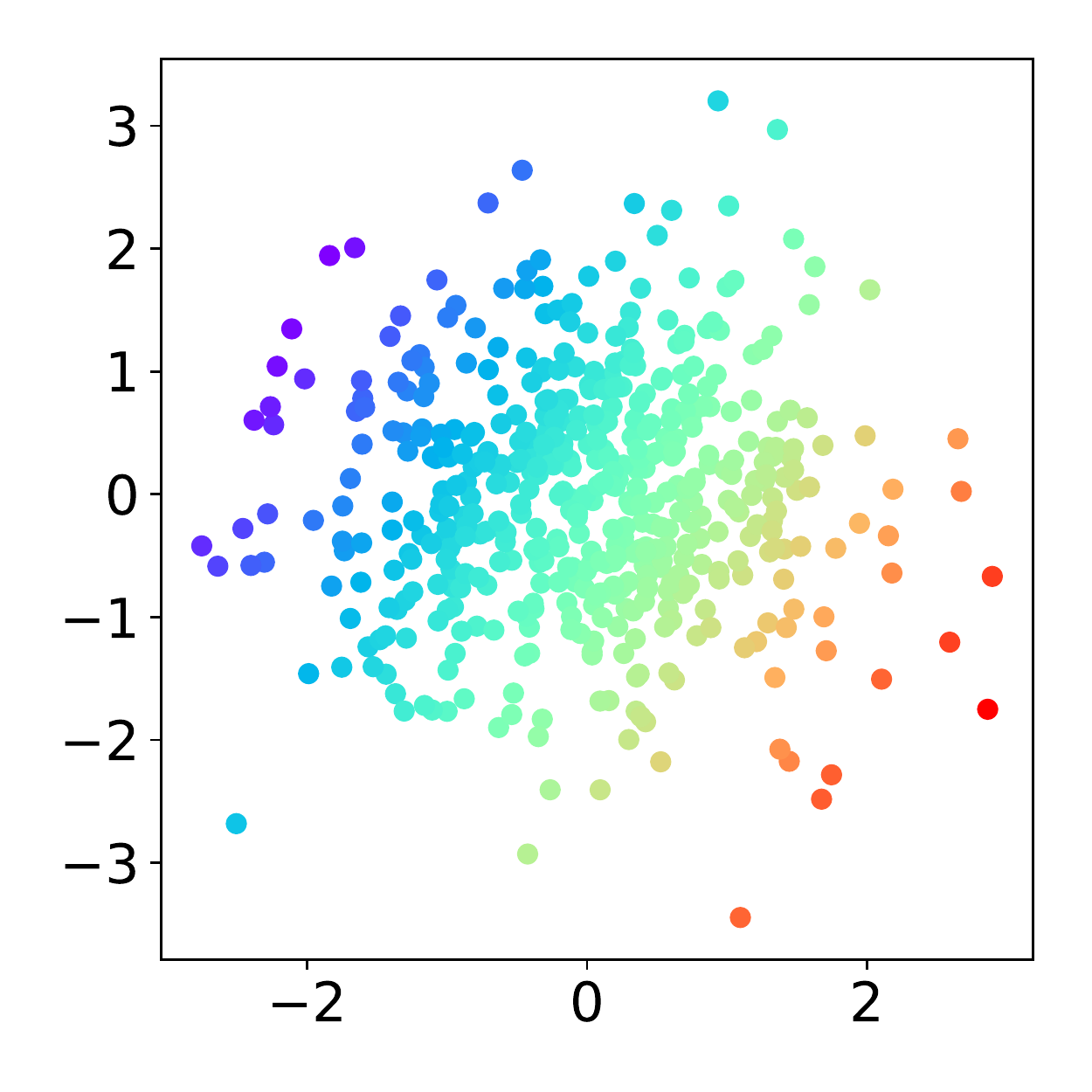}
  \caption{}
  \label{fig:sub-fifth}
\end{subfigure}
\begin{subfigure}{.32\linewidth}
  \centering
  \includegraphics[width=0.75\linewidth]{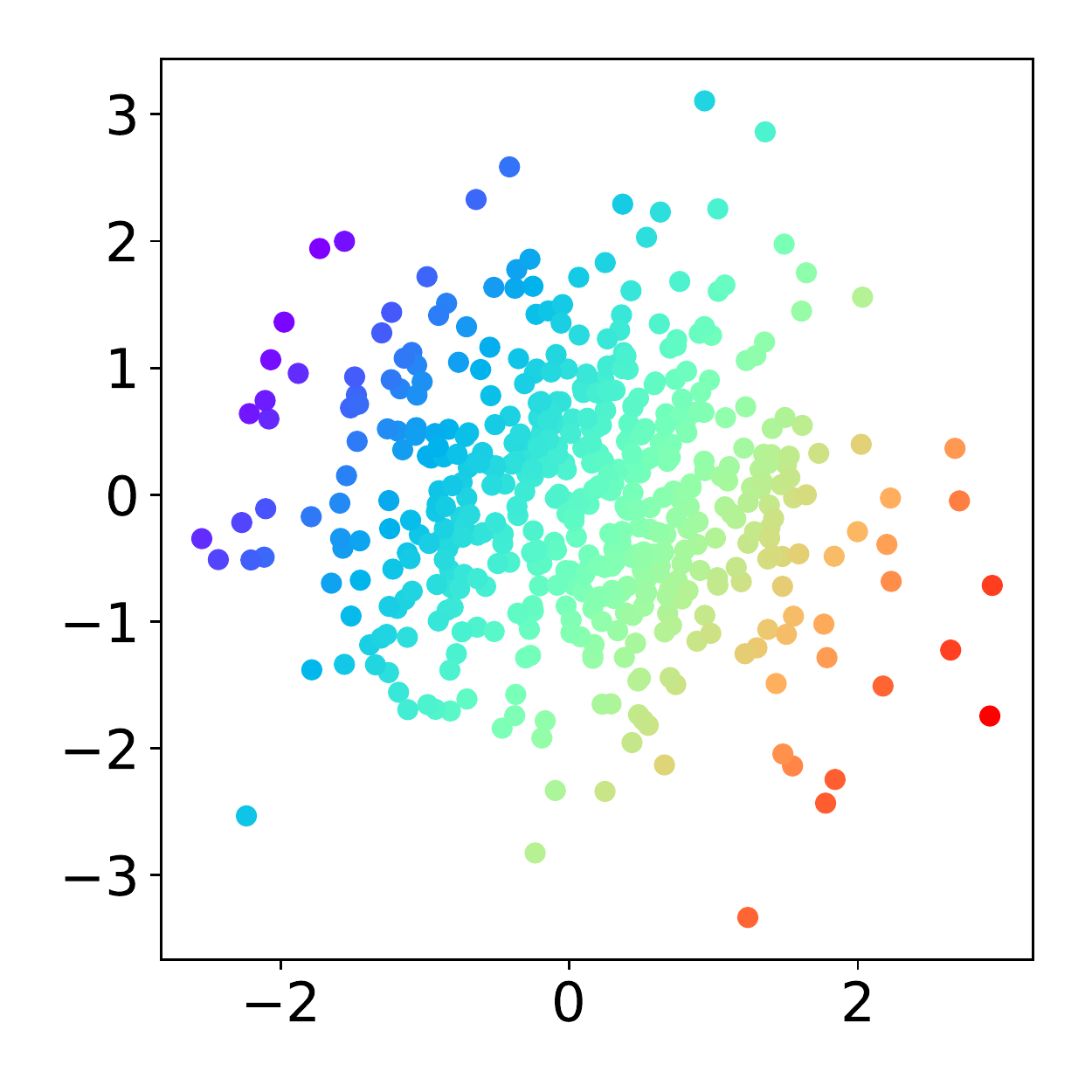}
  \caption{}
  \label{fig:sub-sixth}
\end{subfigure}
\caption{(a) Original data points (colored according to x-axis), Data points after being transported by: (b) WT, (c) IAF \cite{IAF}, (d) CP Flow \cite{ConvexPotentialFlows} (e) \cmgn*, (f)\ \mmgn*. *Our methods.}
\label{fig:coupling_figs}
\end{figure}

We consider the Monge problem in Eqn.~\ref{eqn:monge} with Gaussian data distribution $p_X$, standard Gaussian prior $p_Y$, and Euclidean transport cost $c$. For this restricted Gaussian case, the Wasserstein distance metric gives a closed form solution for the optimal cost. We study cases where the data samples lie in $d=2$ and $16$ dimensions, as in \cite{ConvexPotentialFlows}, and compare our results with Convex Potential Flows (CP Flows) \cite{ConvexPotentialFlows}, Invertible Autoregressive Flows (IAFs) \cite{IAF} and Cholesky-based whitening transform (WT). All models are trained to minimize KL-divergence with the standard normal prior distribution and rely on the change-of-variable formula for probability densities, as discussed in \cite{ConvexPotentialFlows}. For results summarized in Table~\ref{tab:coupling_results}, a lower negative log-likelihood (NLL) signifies that model outputs better fit the standard normal distribution; a lower cost means that data points are moved less when transforming them to fit the standard normal prior. We observe that for $d=2$, both \cmgn\; and \mmgn\; achieve a lower NLL than CP Flow and IAF. Furthermore, Fig.~\ref{fig:coupling_figs} shows how our methods reduce the original data points' movement and incur the smallest transport cost. Our methods are competitive with the others when $d=16$, as they achieve lower NLL than CP Flow and a significantly lower cost than IAF.
\begin{table}[htbp]
\centering
\ninept
\caption{Gaussian optimal coupling results (lower values are better). The optimal costs are $d=2$ and $d=16$ are $3.71$; and $211.97$ respectively. *Our methods.}
\begin{tabular}{ccccc}
\cmidrule(l){2-5}
 & \multicolumn{2}{c}{$d=2$} & \multicolumn{2}{c}{$d=16$} \\
 & NLL & Cost & NLL & Cost \\
 \cmidrule(l){2-5}
Whitening Transform & 2.83 & 4.13 & 22.70 & 235.28 \\
IAF \cite{IAF} & 3.00 & 4.04 & 22.53 & 234.82 \\
CP Flow \cite{ConvexPotentialFlows} & 3.10 & 3.47 & 22.71 & 210.10 \\
\cmgn* & 2.86 & 3.30 & 22.67 & 212.66 \\
\mmgn* & 2.87 & 3.21 & 22.61 & 211.84 \\ \cmidrule(l){2-5}
\end{tabular}
\label{tab:coupling_results}
\end{table}

\subsection{Color Domain Adaptation for Autonomous Driving}
\begin{figure}[tbp]
\centering
\ninept
\begin{subfigure}{\linewidth}
  \centering
  \includegraphics[width=0.24\linewidth]{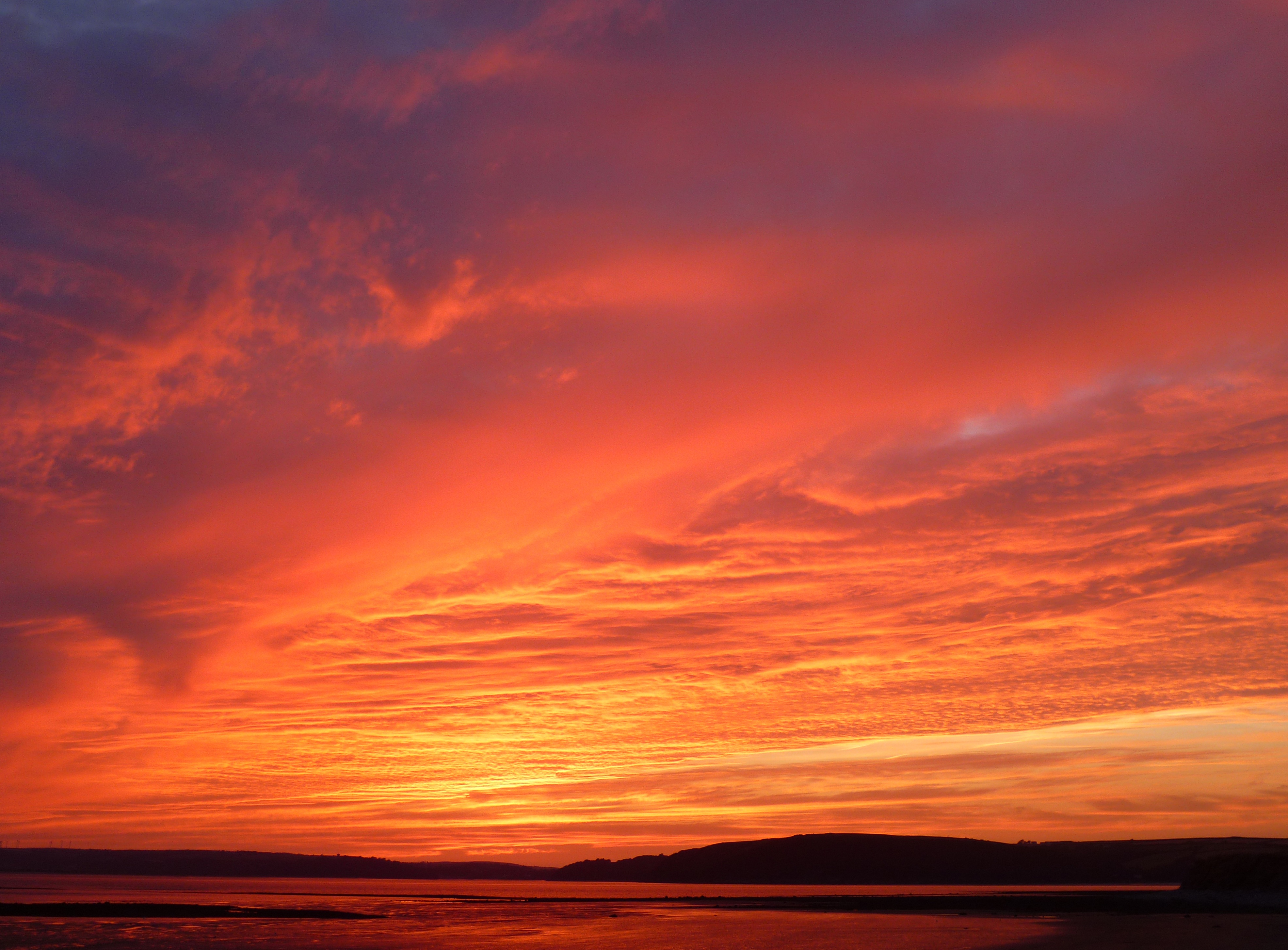}
  \caption{Target}
  \label{fig:sub-first3}
\end{subfigure}
\newline
\begin{subfigure}{.24\linewidth}
  \centering
  \includegraphics[width=\linewidth]{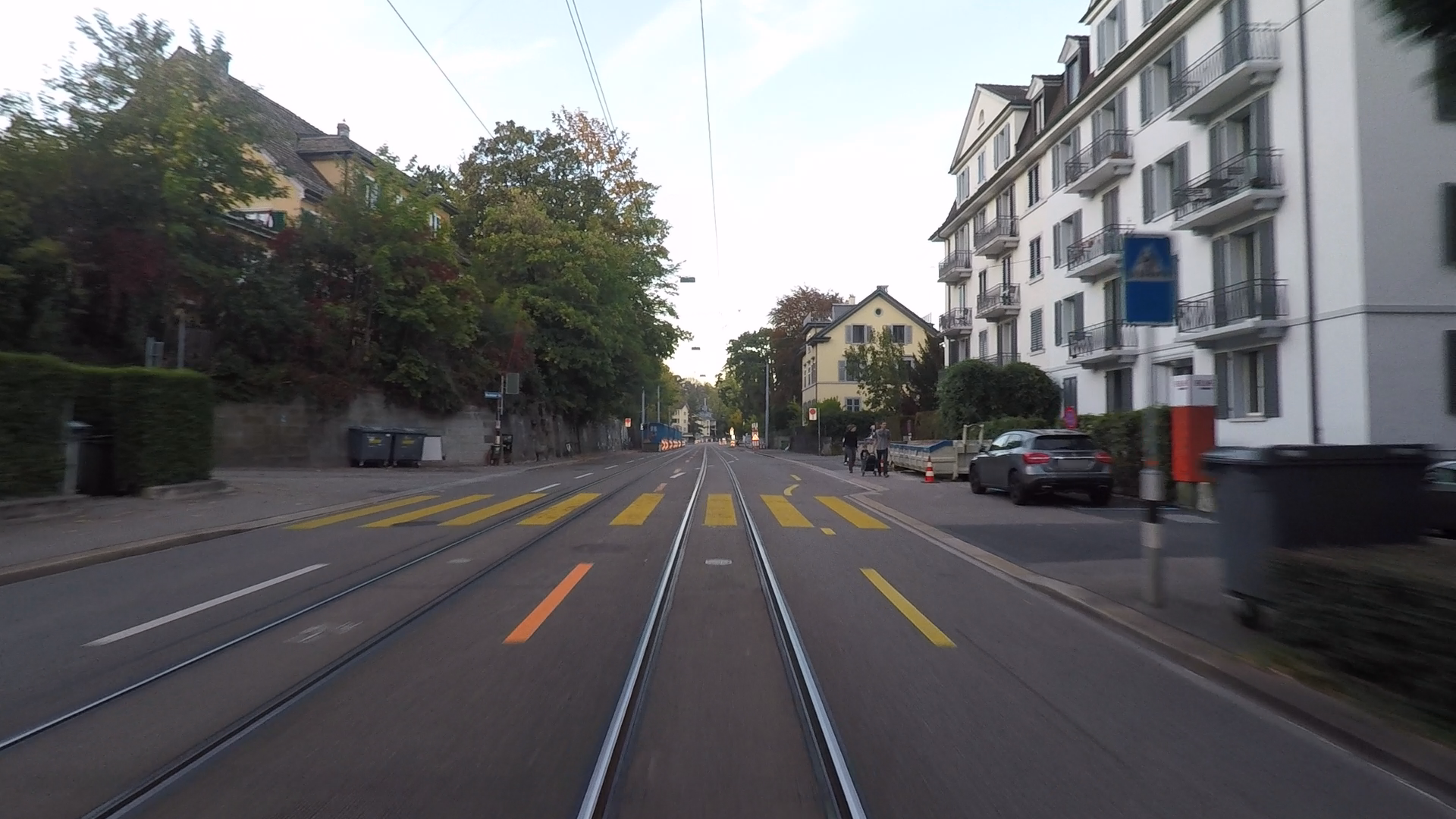}
  \label{fig:sub-second3}
\end{subfigure}
\begin{subfigure}{.24\linewidth}
  \centering
  \includegraphics[width=\linewidth]{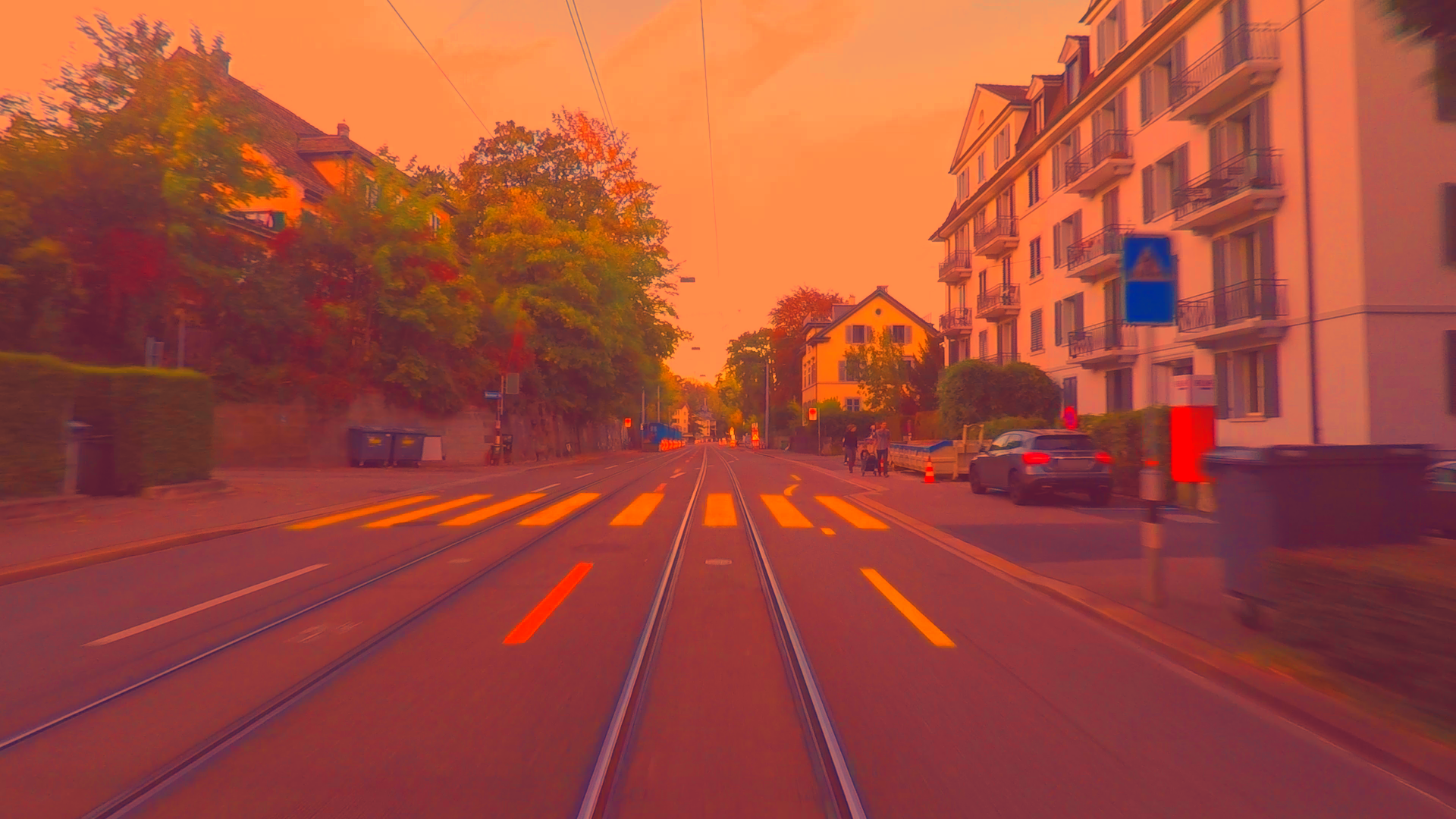}
  \label{fig:sub-third3}
\end{subfigure}
\begin{subfigure}{.24\linewidth}
  \centering
  \includegraphics[width=\linewidth]{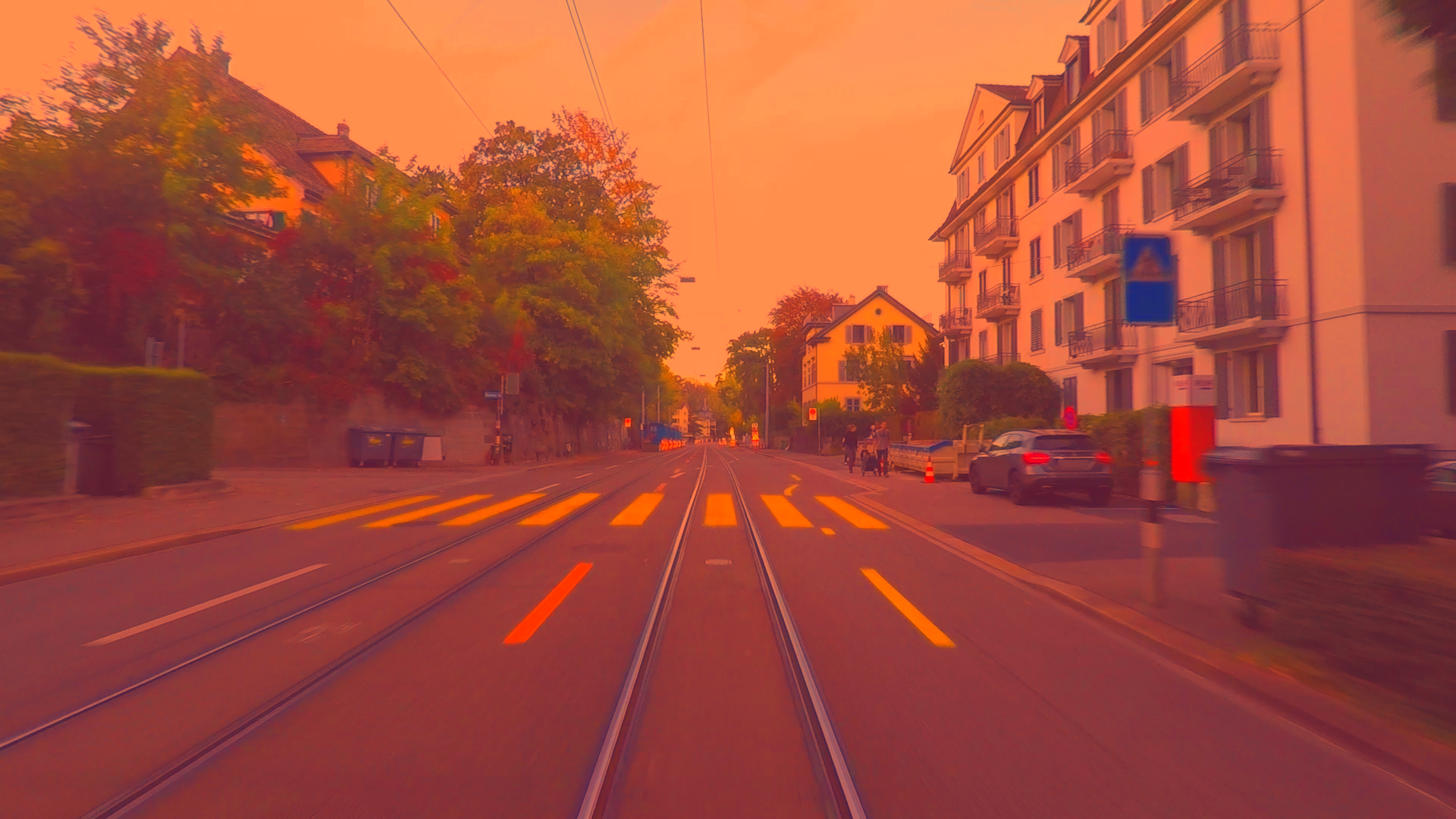}
  \label{fig:sub-fourth3}
\end{subfigure}
\begin{subfigure}{.24\linewidth}
  \centering
  \includegraphics[width=\linewidth]{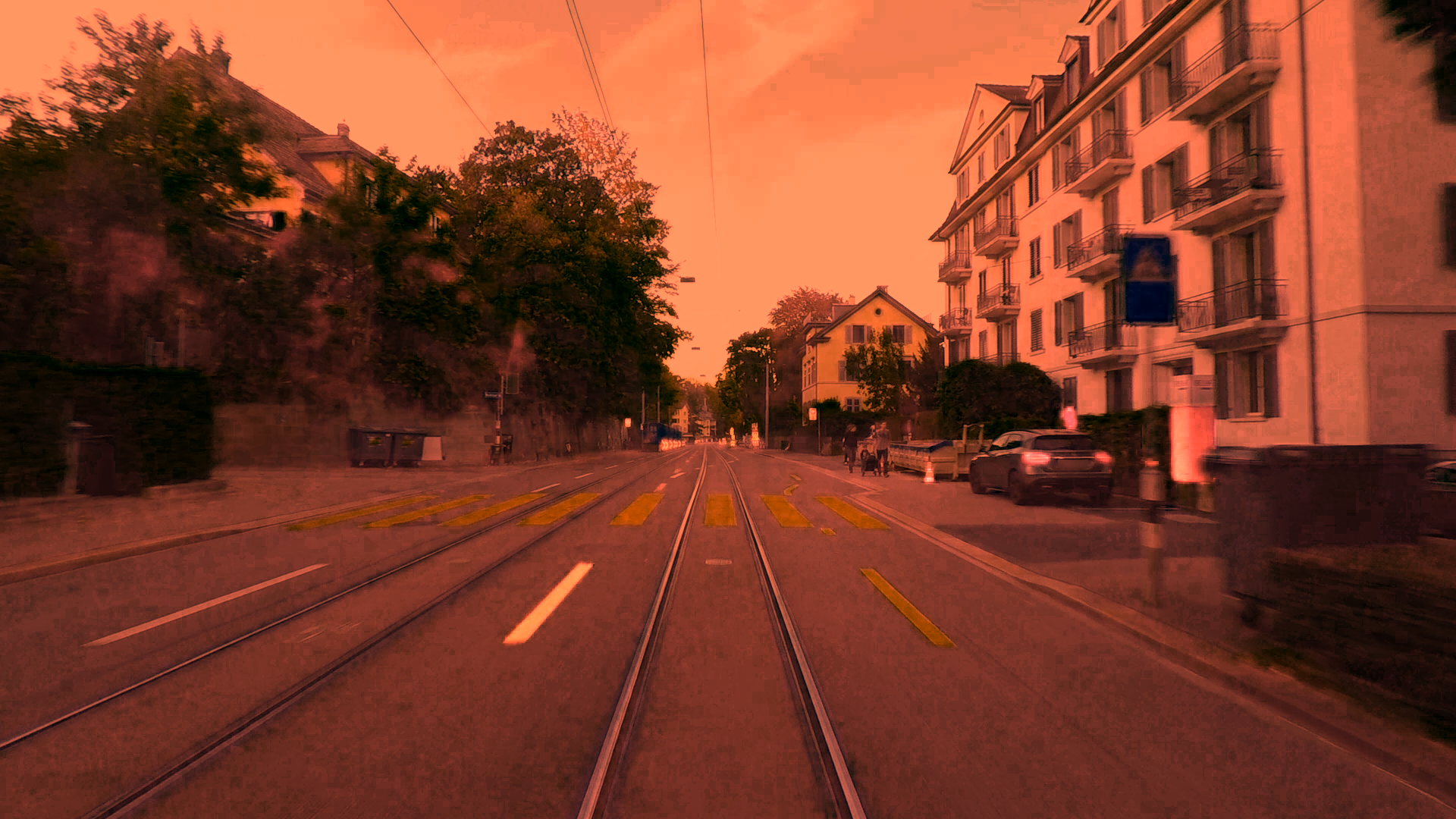}
  \label{fig:sub-fifth3}
\end{subfigure}
\newline
\begin{subfigure}{.24\linewidth}
  \centering
  \includegraphics[width=\linewidth]{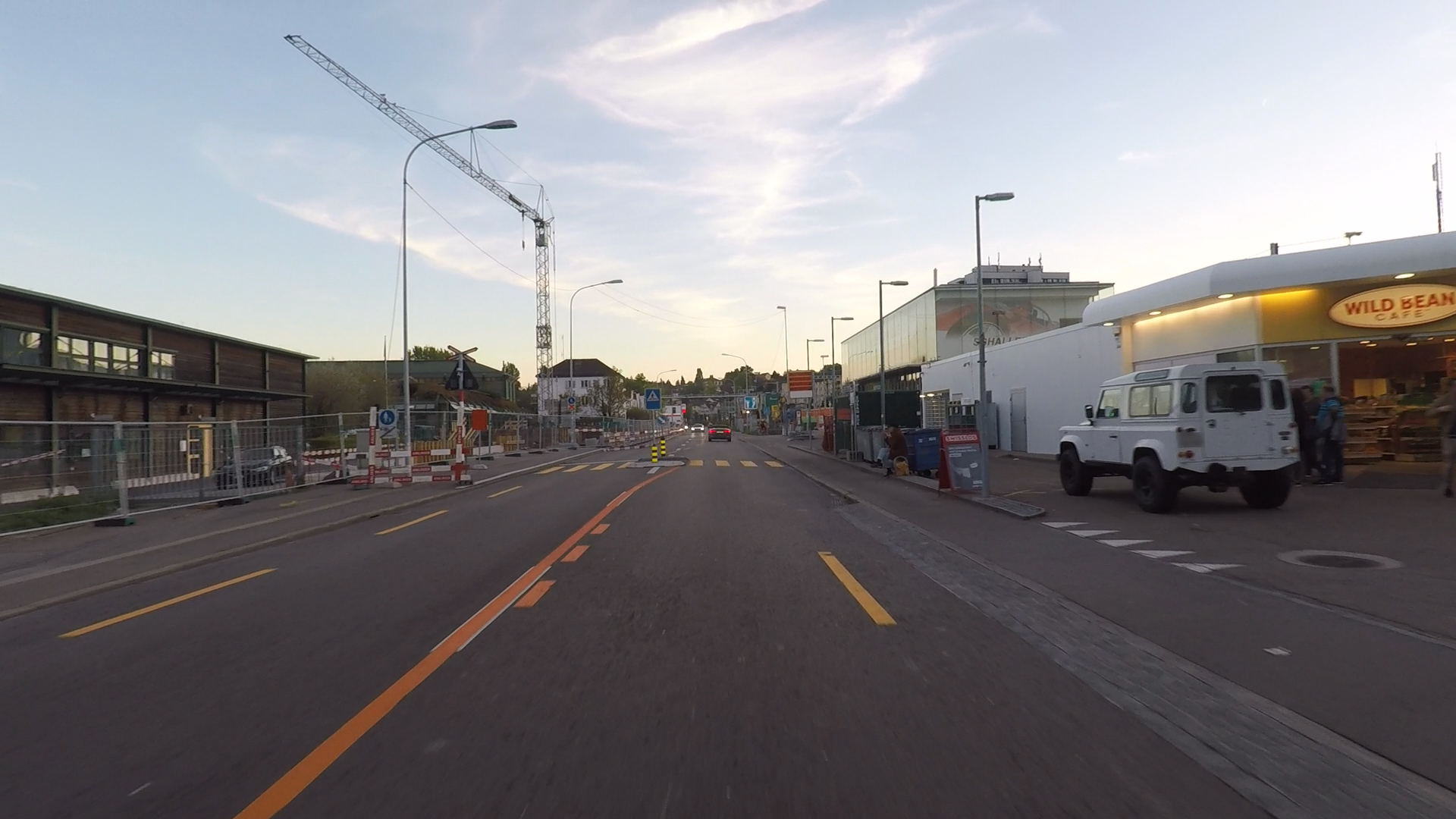}
  \caption{Original}
  \label{fig:sub-sixth3}
\end{subfigure}
\begin{subfigure}{.24\linewidth}
  \centering
  \includegraphics[width=\linewidth]{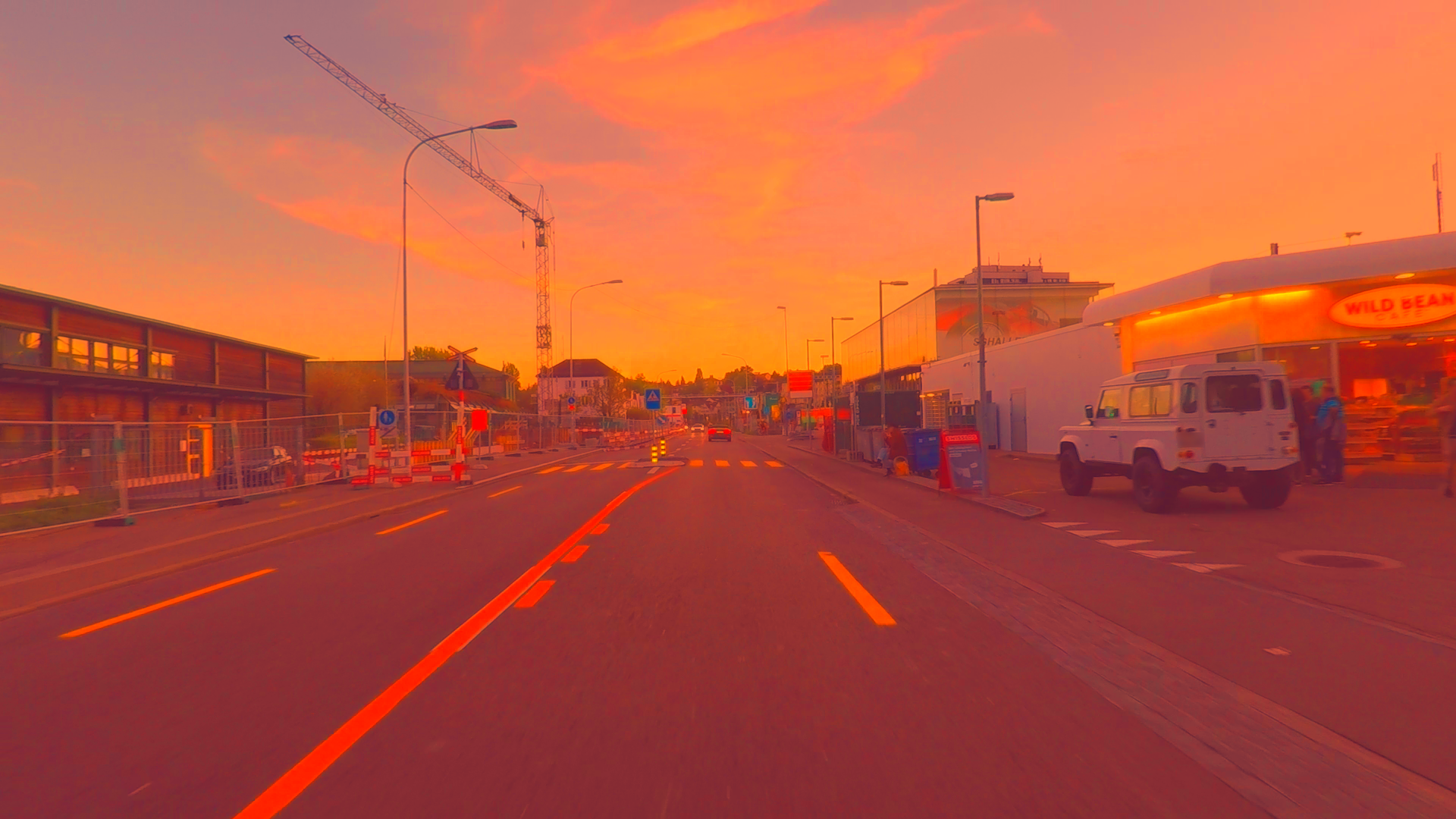}
  \caption{\cmgn*}
  \label{fig:sub-seventh3}
\end{subfigure}
\begin{subfigure}{.24\linewidth}
  \centering
  \includegraphics[width=\linewidth]{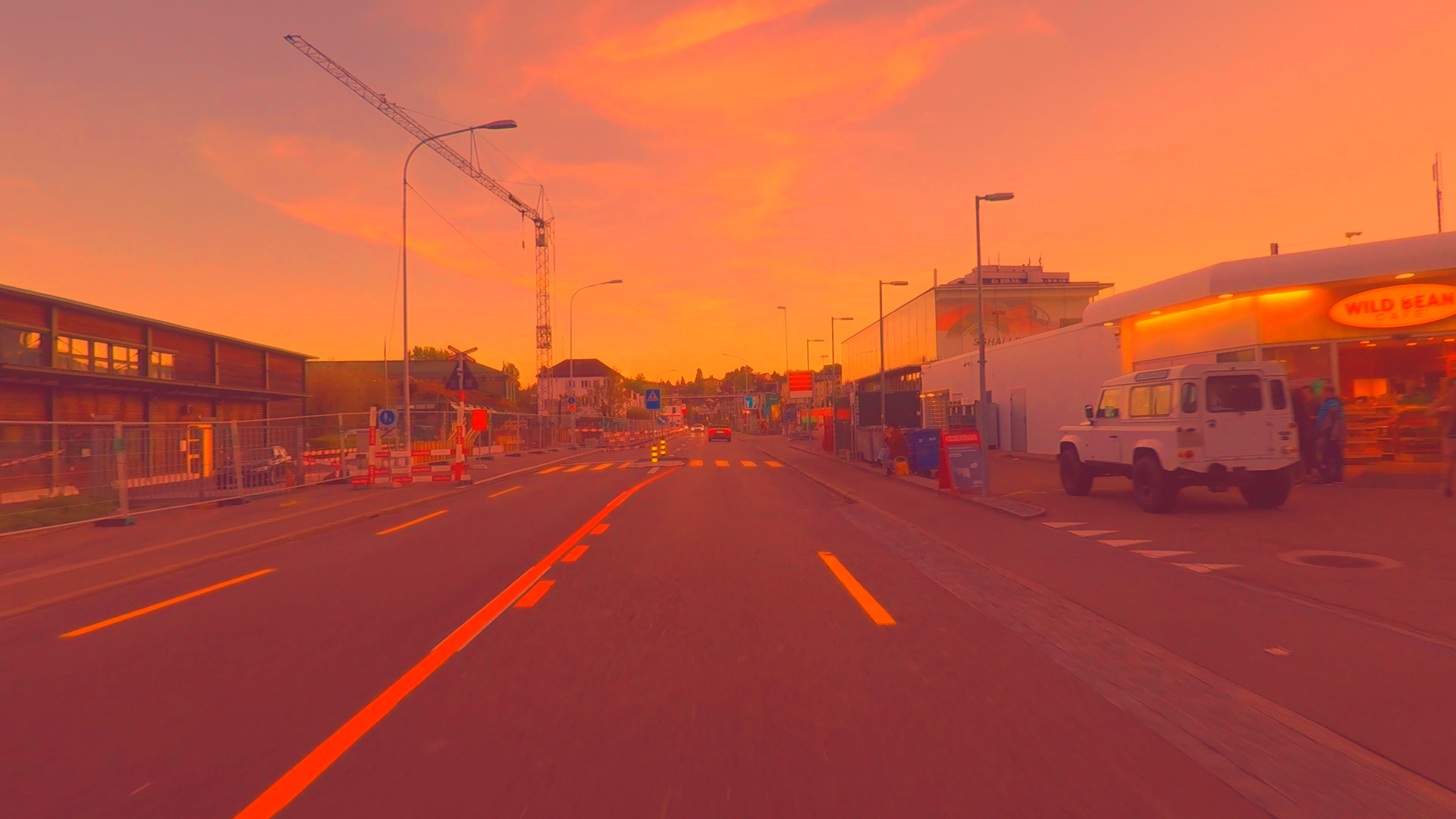}
  \caption{\mmgn*}
  \label{fig:sub-eighth3}
\end{subfigure}
\begin{subfigure}{.24\linewidth}
  \centering
  \includegraphics[width=\linewidth]{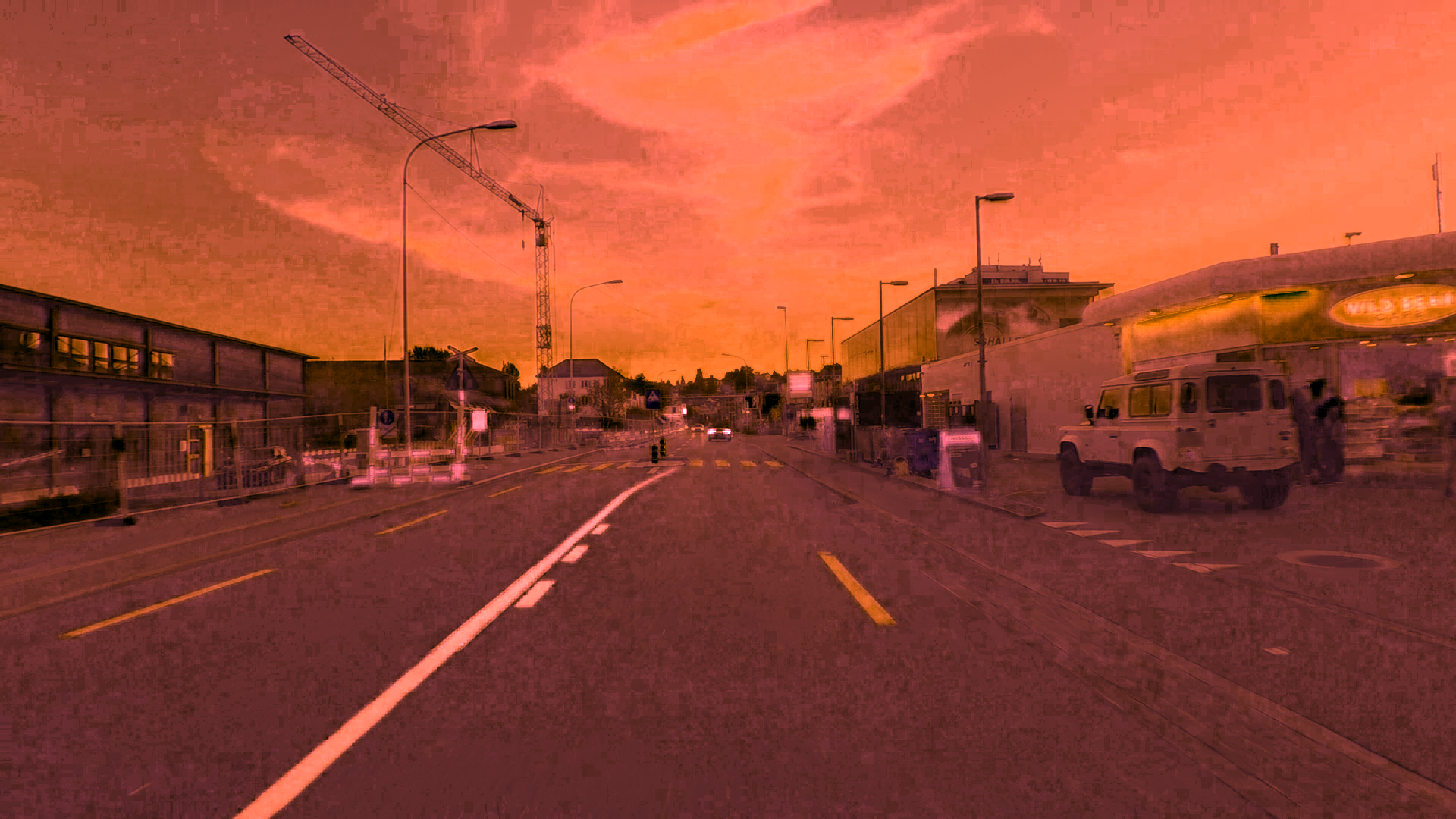}
  \caption{OTKer \cite{perrot2016mapping}}
  \label{fig:sub-ninth3}
\end{subfigure}
\caption{ Color domain adaptation results. Top image: target sunset image.  First row: the single training image. Second row: an example test image. *Our methods.}
\label{fig:color_figs}
\end{figure}

We trained \cmgn\ and \mmgn\ to map daytime road scenes to identical scenes at sunset time, thereby affordably augmenting existing human-annotated image datasets. First, we fit a multivariate Gaussian distribution to the pixel colors present in a target image. Then, similar to Section 5.2, we train our network to learn a mapping from source training image pixels to the target multivariate Gaussian by minimizing the NLL of the mapped image pixel colors.
We train our models using a single target image of a sunset and a single training image of a road from Dark Zurich \cite{SDV19, SDV20} dataset's validation set.
Fig.~\ref{fig:color_figs} shows that both architectures are able to achieve training and test results comparable to a baseline that solves a convex relaxation of the optimal transport problem with added regularizers and Gaussian kernel functions \cite{perrot2016mapping}.

\section{Conclusion}
\label{sec:conclusion}
We propose two monotone gradient network architectures, \cmgn\; and \mmgn, for learning the gradients of convex functions. To the best of our knowledge, this is the first work to directly parameterize and learn gradients of convex functions, without learning the underlying convex function or its Hessian. We show that our networks are guaranteed to represent the gradient of a convex function, suitable for high-dimensional problems, and straightforward to train. We demonstrate that the proposed architectures can learn monotone gradients more accurately and with far fewer parameters than state of the art methods.

\bibliographystyle{IEEEbib}
\bibliography{refs}

\end{document}